\documentclass{article}

\usepackage{microtype}
\usepackage{graphicx}
\usepackage{amsmath}
\usepackage{booktabs} % for professional tables
\usepackage{hyperref}
\usepackage{todonotes}
\usepackage{numprint}
\usepackage{array,multirow,graphicx}
\usepackage{float}
\usepackage{tabulary}
\DeclareRobustCommand{\g}[1]{{\sethlcolor{green}\hl{\textit{#1}}}}

\definecolor{myred}{RGB}{255, 160, 195}
\definecolor{myblue}{RGB}{86, 201, 255}
\DeclareRobustCommand{\r}[1]{{\sethlcolor{myred}\underline{\hl{#1}}}}

\usepackage{color,soul}
\usepackage[accepted]{icml2020}
\usepackage{url}

\usepackage{breakurl}

\usepackage{amsmath,amsfonts,amsthm,amscd,amssymb}
\DeclareMathOperator*{\argmax}{arg\,max}

\icmltitlerunning{Self-Supervised and Controlled Opinion Summarization}

\begin{document}

\twocolumn[
%\icmltitle{Abstractive Multi-Document Opinion Summarization\\ Through Self-Supervision and Control}
%\icmltitle{Abstractive Opinion Summarization\\ Through Self-Supervision and Control}
%\icmltitle{Self-Supervised \& Controlled Abstractive Summarization of Reviews}
%\icmltitle{Self-Supervised Abstractive Opinion Summarization}
\icmltitle{Self-Supervised and Controlled \\ Multi-Document Opinion Summarization}

\icmlsetsymbol{equal}{*}
\begin{icmlauthorlist}
\icmlauthor{Hady Elsahar}{naver}
\icmlauthor{Maximin Coavoux}{uga}
\icmlauthor{Matthias Gall\'e}{naver}
\icmlauthor{Jos Rozen}{naver}
\end{icmlauthorlist}

\icmlaffiliation{uga}{Univ. Grenoble Alpes, CNRS, Grenoble INP, LIG, France}
\icmlaffiliation{naver}{Naver Labs Europe, Meylan, France}
\icmlcorrespondingauthor{Hady Elsahar}{hady.elsahar@naverlabs.com}
\icmlkeywords{self-supervised, summarization, control tokens}
\vskip 0.3in
]
\printAffiliationsAndNotice{} % otherwise use the standard text.

\begin{abstract}

We address the problem of unsupervised abstractive summarization of collections of user generated reviews through self-supervision and control.
We propose a self-supervised setup that considers an individual document as a target summary for a set of similar documents.
This setting makes training simpler than previous approaches by relying only on standard log-likelihood loss.

We address the problem of hallucinations through the use of control codes, to steer the generation towards more coherent and relevant summaries.
Finally, we extend the Transformer architecture to allow for multiple reviews as input. 

Our benchmarks on two datasets against graph-based and recent neural abstractive unsupervised models show that our proposed method generates summaries with a superior quality and relevance.
This is confirmed in our human evaluation which focuses explicitly on the faithfulness of generated summaries
We also provide an ablation study, which shows the importance of the control setup in controlling hallucinations and achieve high sentiment and topic alignment of the summaries with the input reviews.

\end{abstract}

\begin{figure*}%
    \centering
    \includegraphics[width=\linewidth]{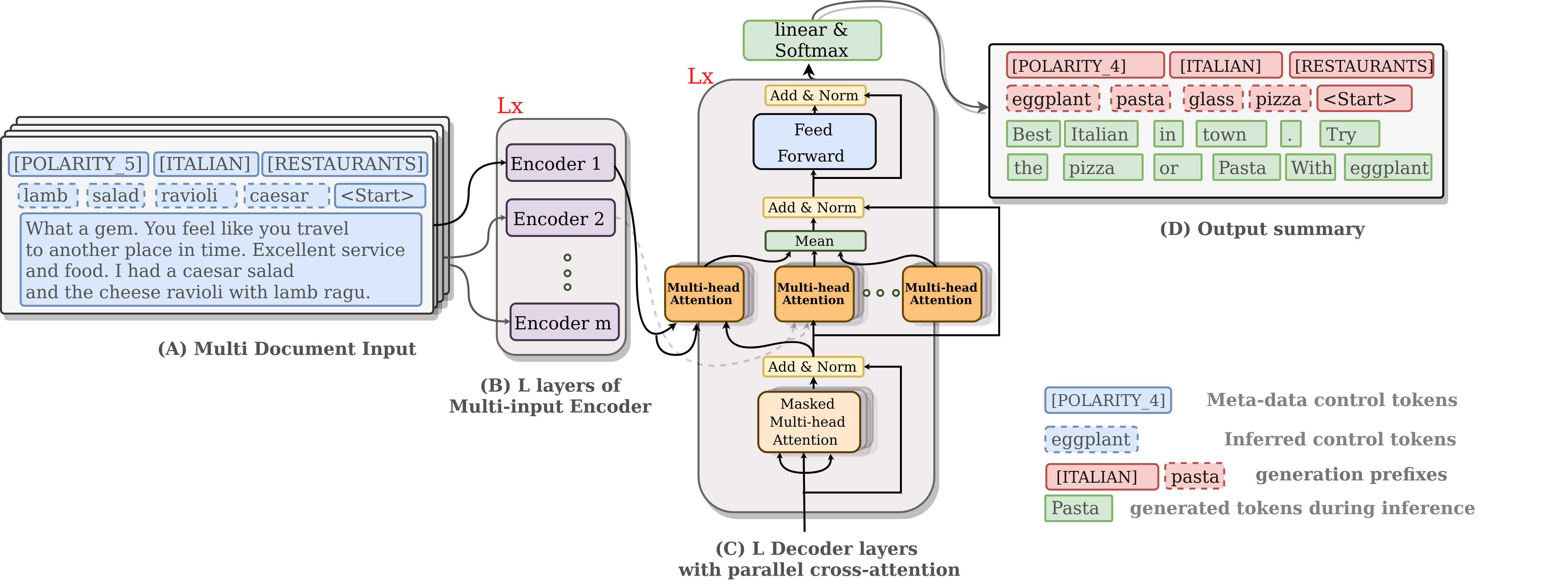} 
    \caption{\small{Description of our proposed model: (A) is the set of input reviews, augmented with control tokens (from meta-data in uppercase, inferred in lowercase). (B) is the encoder, which is run separately on each input review. The standard Transformer decoder is modified in (C) to allow for Parallel cross-attention on different inputs separately. Finally, (D) is the generated output. During inference the control tokens are fed as prompts to the decoder and generation starts afterwards.}}
    \label{fig:main_figure}%
\end{figure*}

\section{Introduction}
\label{sec:intro}

Recent progress in unsupervised methods has created breakthroughs in natural language processing applications, such as machine translation \cite{ArtetxeLAC18,LampleCDR18}.
Those have been mostly based on a bootstrapping approach, which consists in iteratively alternating between two representations, and optimizing a reconstruction loss.
Machine translation is the most successful of those applications, but other applications include Question-Answering \citep{lewis2019} and parsing~\citep{drozdov2019}.
While similar ideas have been applied as well for video summarization \citep{yuan2019}, such a bootstrapping approach seems less suited for summarization, because of the inherent information loss when going from the full text to the summarized one.
Existing unsupervised approaches for summarization therefore relied mostly on extractive graph-based systems~\citep{mihalcea-tarau-2004-textrank}.
Only recently have there been proposals for unsupervised abstractive summarization, using auto-encoders~\citep{meansum,brazinskas_unsupervised_2019}.
However, these set-ups are quite complex, requiring a combination of loss functions \citep{meansum} or hierarchical latent variables~\citep{brazinskas_unsupervised_2019} to ensure that the generated summaries remain on-topic.

In this paper, we investigate a self-supervised approach for multi-document opinion summarization.
In this setting, there are multiple opinions (reviews), one entity (products, venues, movies, etc) and the goal is to extract a short summary of those opinions.
Our approach is based on self-supervision and does not require any gold summaries.
We train a supervised model on examples artificially created by selecting (i) one review that will act as a target summary and (ii) a subset of reviews of the same entity that acts as a document collection.

Neural models have a known problem of hallucination \cite{rohrbach-etal-2018-object}, which can be utmost misleading in natural language generation tasks as the fluency of those models often distract from the wrong facts stated in the generated text.
To reduce this effect, we propose to use control tokens \citep{fan2017,keksar_ctrl_19}.
Control tokens are discrete variables that are used to condition the generation. 
Different from previous work, our goal is not to allow users to control the generated text, but instead to steer the generated text to produce an output which is consistent with the input documents to be summarized. 

Our main contributions are therefore three-fold:
\begin{itemize}
    \item performing multi-document summarization by modelling it as a self-supervised problem where one document acts as the summary of a subset. 
    We carefully select those two, and link the resulting formulation to a recently proposed theoretical framework \cite{peyrard_simple_2019} (Sect.~\ref{sect:selfsuper});
    \item using control tokens to steer the model towards consistency, increasing relevance of the generated summary  (Sect.~\ref{sect:controlcodes});
    \item an application of multi-input transformer model~\citep{LibovickyHM18} to summarization. This model encodes each input independently, and at decoding time applies parallel attention to each encoded input (Sect.~\ref{sect:transformer-model}).
\end{itemize}

Our experimental results (Sect.~\ref{sect:experiments} and \ref{sect:results}) show that our proposed approach outperforms existing models on two datasets: Yelp reviews on venues \citep{meansum} and Rotten Tomatoes movie reviews \citep{wang2016}.
We focus the human evaluation on the faithfulness of the generated reviews and they confirm that the generated summaries are more factually correct than the compared baseline.

\section{Related Work}

\paragraph{Unsupervised Opinion Summarization}
Unsupervised Multi-Document summarization methods encompass both \textit{extractive} and \textit{abstractive} approaches.
Extractive summarization consists in selecting a few sentences from the input documents to form the output summary.
\citet{RADEV2004919} proposed to rank sentences according to their relevance to the whole input, representing sentences as tfidf bags of words and the input as the centroid vector of its sentences.
Recent refinements of this approach include using distributed word representations~\cite{rossiello-etal-2017-centroid} or ranking whole summaries instead of individual sentences \cite{gholipour-ghalandari-2017-revisiting}.
Graph-based methods, such as LexRank \cite{Erkan:2004:LGL:1622487.1622501} or TextRank \cite{mihalcea-tarau-2004-textrank,zheng-lapata-2019-sentence}, work by constructing a graph whose nodes are the sentences from the input documents and whose edges indicate a high word overlap between two sentences. 
Then, they use the PageRank algorithm to extract the sentences with the highest centrality.
In contrast to these methods, we focus on abstractive summarization methods.

Abstractive methods for summarization are in principle able to generate new words and sentences that do not occur in the input documents and therefore produce more fluent text.
Non-neural abstractive methods \cite{ganesan-etal-2010-opinosis,nayeem-etal-2018-abstractive} are also graph-based, but construct graphs whose nodes are word types and edges indicate the immediate precedence relationship between two instantiations of the word type in a sentence.
The summary is extracted by finding salient paths in the graph.
%Instead, our system is based on neural NLG.

Recently, a few approaches for \textit{neural} unsupervised abstractive summarization have been proposed.
\citet[MeanSum]{meansum} introduced a summarization system based on a review autoencoder.
At inference time, MeanSum encodes every review for a product to a vector, computes the centroid of reviews' vectors and uses this centroid to seed the decoder and generate a summary.
However, averaging representations of statements that are sometimes contradictory tends to confuse the decoder, and to lead it to rely on only language modeling for generating the output summary,  thus ignoring the input signal.
To deal with this limitation,  \citet{coavoux-etal-2019-unsupervised} proposed to add a clustering step to identify similar reviews and to generate one sentence per such found cluster: the averaging step only targets similar reviews.
Contemporaneous to this work, \citet{brazinskas_unsupervised_2019} proposed to solve the problem of unsupervised summarization of reviews through an auto-encoder with latent variables.
Their proposed way of solving the problem of hallucinating content from other categories is to use one latent variable per product, and let the decoder access all the reviews of a product.
Compared to it, we argue that our self-supervised setting is simpler as it relies on training with standard cross-entropy.
In addition, the use of Transformer (as opposed to GRU in their case) makes it possible to apply separate attentions to each input.

A similar idea was very recently proposed for pretraining summarization models.
\citet{Pegasus} masks out full sentences from a document, and trains a model that predicts those sentence from the surrounding text.
Our self-supervision training mechanism can be seen as a multi-document version of that.

% ~\cite{HuYLSX17} --> VAE for controlled generation?

\citet{west-etal-2019-bottlesum} introduced a self-supervised system for sentence compression: they design an unsupervised extractive system and use it to generate data to train a supervised neural sentence compressor. However, their two-level system works at the level of single sentences whereas our end-to-end approach summarizes sets of reviews with multiple sentences.

%\cite{peyrard_simple_2019}
%\cite{angelidis-lapata-2018-summarizing}

\paragraph{Controlled Generation}
% RL for control 
We rely on controlled natural language generation to steer the generation away from hallucinations. %
Controllable text generation has been previously investigated to apply global constraints on text generation. 
Previous work proposed fine-tuning NLG models to provide control. To allow back-propagation through the discrete sampling process of text generation several proposals have used policy gradient methods, most notably REINFORCE~\cite{williams1992reinforce} for applications such as machine translation~\cite{RanzatoCAZ15_mt_reinforce,wu_reinforce_mt}, image-to-text generation~\cite{liu2017_img_reinforce}, dialogue generation~\cite{LiMRJGG16_dialogue_reinforce} and visual question answering~\cite{Yi0G0KT18_VQA_reinforce}.
Other work has relied on continuous approximation methods, most notably Gumbel-Softmax~\cite{JangGP17_original_gumbel} such as in~\citet{meansum,YangHDXB18_yang_gumbel}.

Other methods of control applied control only at inference time. 
Weighted decoding, introduced by \citet{holtzman-etal-2018-learning}, was shown to be challenging, and to often lead to sacrificing fluency and coherence~\citep{SeeRKW19}.
Constrained beam search~\cite{AndersonFJG17_img_beamsearch,HokampL17_MT_img_beamsearch,PostV18_beamsearch_img_MT} is slower, requires in practice very large beam sizes, and does not enable soft constraints. Finally,  updating the decoder hidden states~\cite{ChenLCB18,dathathri2019plug} requires an extra training step.

The first introduction of control codes to neural generation models has been an early form of copy mechanism to overcome the rare word problem~\cite{LuongSLVZ15_rare,ElSaharGL18} and  has recently shown a wide adoption - due to its simplicity and effectiveness - to steer large scale language models toward desired traits such as general aspects \citep{keksar_ctrl_19} or structured fields \citep{grover}.
Previous work for controlling large-scale language models has relied on a predefined set of bag of control tokens, collected either manually~\cite{keksar_ctrl_19} or from dictionaries~\cite{dathathri2019plug}, which can lead to low domain coverage. %
Regularized classification models have intrinsic feature selection capabilities~\cite{ng2004}, that have been exploited before for lexicon generation from sentiment classifiers~\cite{labr2014,ElSaharE15}.
These approaches generate more relevant lexicons than traditional topic models such as LDA~\cite{blei2003}. In this work, to automatically generate bag of control tokens we follow the same approach which does only rely on the category meta-data provided with the reviews. In the absence of such meta-data information several approaches have relied instead creating seed lexicons using unsupervised or weakly supervised aspect extractors ~\cite{he-etal-2017-unsupervised,angelidis-lapata-2018-summarizing}

\paragraph{Hierarchical encoding.} In order to allow a neural summarizer to read several sections \citet{cohan2018} proposes a hierarchical LSTM that works at two level.
Most similar to our proposed method, \citet{liu-lapata2019} extends a Transformer network to read several ranked paragraphs as input, avoiding a retrieve-then-read pipeline.
In multi-document summarization the paragraphs are not ranked but independent. 
This entails a significant change model-wise, and we propose to encode each review independently (avoiding their inter-paragraph self-attention) and only adapt the decoder-encoder attention.

%%%%%% SELF SUPERVISION %%%%%%
\section{Self-Supervision}
\label{sect:selfsuper}
In order to create our training dataset we assume that a review $s_i$ for an entity (venue or product) can serve as a summary for a set of other similar reviews $D_i$. 
This simple intuition allows us to create training points $(D_i,s_i)$ in a very similar way to what the model will experience at inference time.
However, there are two issues with this approach.
First, the potential set of training points is too large to explore exhaustively.
Given the set of all reviews $\mathcal{D}$ the total number of possible input-output pairs is $2^{|\mathcal{D}|-1} \times |\mathcal{D}|$.
Second, the assumption that any review is fit to serve as a summary for any set of other reviews is obviously not true, and might yield a very noisy training dataset.

To solve the combinatorial explosion, we limit the size of~$D_i$ to~$k$, and from a given $s_i$, we look for a set of $k$ \textit{good} reviews~$D_i$, for which $s_i$ serves as a good summary.
Fixing $k$ also simplifies training, and enables comparison with previous work where the number of input reviews is fixed~\cite{meansum,brazinskas_unsupervised_2019}.
Both $s_i$ and all members of $D_i$ are reviews of the same entity.

Having $s_i$ fixed, we now search for reviews $d_1, \dots, d_k$ for which $s_i$ is a relevant review:
\begin{equation} \label{eq:relevance}
\begin{split}
        \textit{rel}(s_i) &= \{d_1, d_2,...,d_k\}, \\
      &= \argmax_{D_i \subset \mathcal{D} \setminus \{s_i\}, |D_i|=k} \; \;  \sum_{d_j \in D_{i}} \text{sim}(s_i, d_j).
\end{split}
\end{equation}

Note that fixing first the target summaries turns traditional approaches upside down.
In particular, a recently proposed theoretical model of importance in summarization \citep{peyrard_simple_2019} defines the importance of a summary based on three aspects: (i) minimum redundancy, (ii) maximum relevance with the input document, and (iii) maximum informativeness. 
In that line of work $D_i$ is considered fixed: redundancy and informativeness are not dependent on $D_i$ and can therefore be ignored when $s_i$ is fixed.
In this setting \citet{peyrard_simple_2019} reduces then to Eq.~\ref{eq:relevance}

Then, we sort the data-points $(d_i, \textit{rel}(d_i))$ according to the value of the relevance ($\sum_{d_j \in \text{rel}(d_i)} \text{sim}(d_i, d_j)$).
Depending on the desired size of the target dataset, we keep the top-$T$ pairs for training.
Limiting $T$ inherently increases informativeness, since it limits the creation of training examples where input and outputs are repetitive similar reviews that might be very prominent on corpora level (e.g. ``Great restaurant.").  
In addition to simplicity, this method enables a fast implementation using state-of-the-art nearest neighbour search libraries~\citep{sklearn}.
For all our experiments we defined \textit{sim} to be the cosine similarity over a tf-idf bag-of-word representation~\cite{tfidf}.

\section{Controlling Hallucinations}
\label{sect:controlcodes}

Hallucinations are pieces of generated text that bear no relationship to the text they were conditioned on.
They are likely to happen in our self-supervised setting, due to the noise from the construction of training instances.
This might happen, for instance, if the synthetically created training data contains a variety of contradictory signals, or because certain types of review are overly present (e.g.\ ``great movie'').
The model might default to those very frequent patterns if it finds itself in a unfrequent state during decoding time.

To alleviate the problem of hallucinations, we propose to use \textit{control tokens} that represent desired traits of the output text to steer the generated text towards more input-coherent summaries.

These control tokens are inferred from each review, and used as prompts at inference time. 
We use two types of codes as follows:

\textbf{1) Metadata control tokens.} Those are special tokens that are associated with each input review, and are the capitalized control tokens in Fig.~\ref{fig:main_figure}.
We use two types of metadata that represent (i) the review \textbf{polarity}, a numerical value denoting the average sentiment score of the input reviews; (ii) and \textbf{categorical tokens} representing the type of the entity of the review (e.g. Deli, Beauty\&Spa, Furniture Stores). 
In the case of the unavailability of meta-data labels for all reviews (as in Rotten-Tomatoes dataset), we infer control tokens with the same process, but using categories predicted by a trained classifiers on labeled examples from the same domain.
\smallskip

\textbf{2) Inferred control tokens.}
We follow recent work~\cite{keksar_ctrl_19,dathathri2019plug} that shows that it is preferable to condition NLG models on control tokens that naturally co-occur in text. 
On one side, this allows for better control, and at the same it seems to be more robust when new (previously unseen) control codes are used. Here, we propose to use control codes that represent informative aspects (e.g.\ wine, service, ingredients) that occur in the input reviews text.
However, instead of relying on manually created bag of control tokens for each desired attribute -- which comes with obvious domain coverage limitations -- we propose to infer those control codes from the text corpus.
To do so, we rely on the intrinsic feature selection capabilities of regularized linear classification models.    
For each category~$\ell$ in the meta-data associated with each review we train a linear support vector machine (SVM) classifier~\cite{VapLer63svm}\footnote{We use \texttt{liblinear}~\citep{fan2008liblinear}.} that learns to classify between reviews from this category and negative examples sampled randomly from the rest of the corpus. 
The features of the SVMs are parameterized by the weight vector $\theta_\ell \in \mathcal{R}^d$, where $d$ is the number of features (in our experiments: all unigrams and bigrams present in the corpus).
We used a squared hinge loss with~$L1$ regularization over $\theta_\ell$ -- the latter to increase sparsity and force feature selection \citep{lasso96,ng2004}. %l1_zheng_2004
Finally, we trim the feature list into those who correspond to positive weights and re-normalize the weights. 
The output of this step is a ranked list of $n$-grams that represent the distinctive aspects of each category.

When creating training data for summarization, we enrich each review with the top weighted $n$-grams of their corresponding categories as follows. 
For a given review $d$ about entity~$p$, we consider all $m$ labels of $p$ and use the weights of the corresponding classifiers $\theta_{\ell_i^{(p)}}$ (for each label $\ell_i^{(p)}$ of~$p$).
We only consider those $n$-grams actually occurring in $d$, and keep the top~8 such features.
Note that these features could come from different classifiers, as we consider all $m$ labels.

During training, each review is enriched with its tailored control codes.
In particular, the reviews acting as summary also contain them, and by construction those are $n$-grams present in the text.
At inference time -- when the target side and hence its control codes are not available -- we select the most repeated control tokens from the input side and feed them as a prefix to the decoder before the start of generation.  
There is clearly a risk that the model just learns to copy the control codes it has seen somewhere in the text.
We check whether this is the case in Sect.~\ref{sect:results}.

\begin{figure}%
    \centering
    \includegraphics[width=0.55\columnwidth]{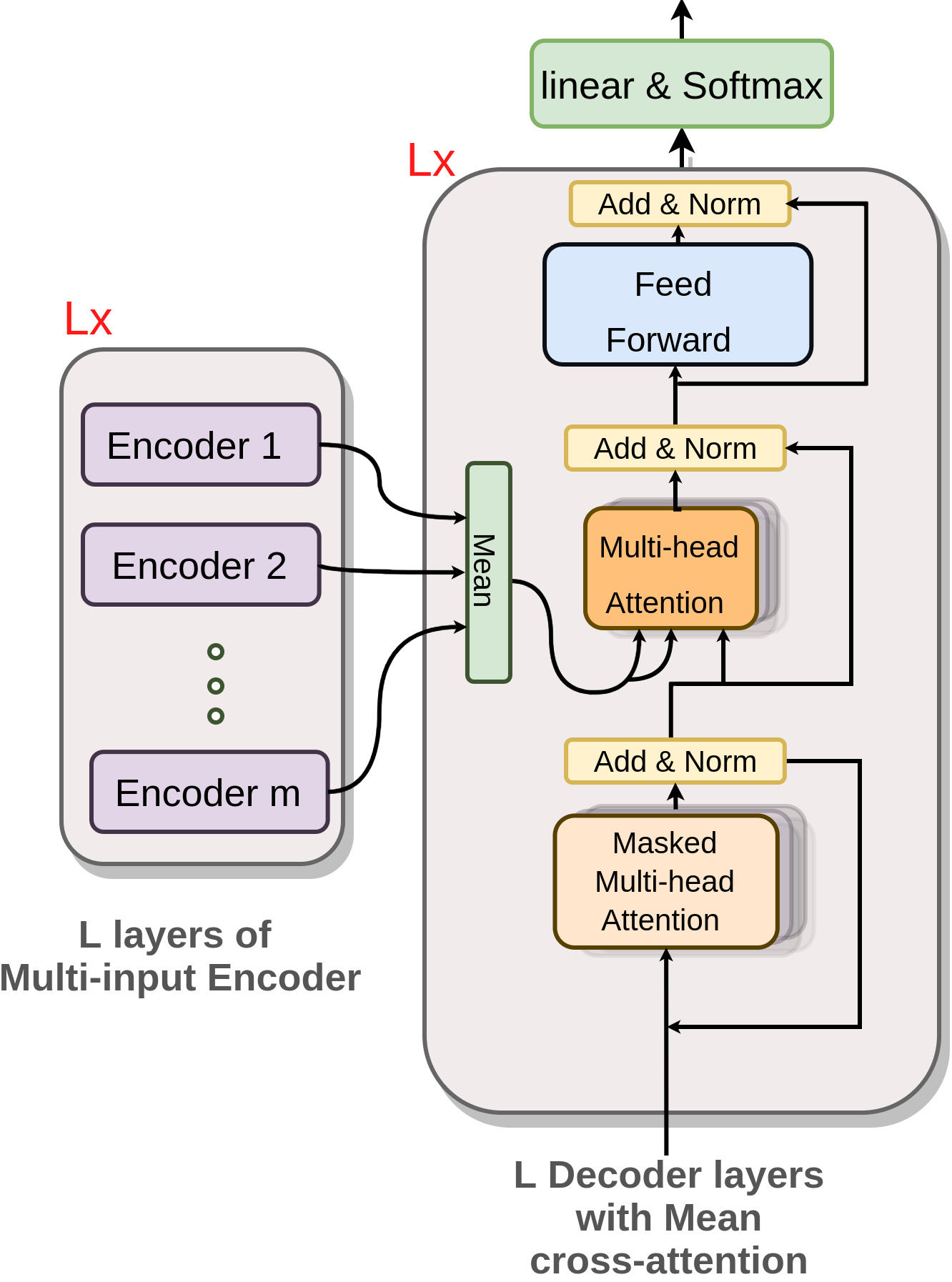} 
    \caption{Figure showing our adaptation of the Transformer cross-attention to allow \textit{Mean} combination of multi-sources.}
    \label{fig:meanattention}%
\end{figure}

%%%%%%%%%%%%%%%%%%%%%%%%%%%%%%%%%%
%%%%%%%%%%%%% MODEL %%%%%%%%%%%%%%
%%%%%%%%%%%%%%%%%%%%%%%%%%%%%%%%%%
\section{Multi-source Transformer Model}
\label{sect:transformer-model}

%% maybe move to intro or related work 
Previous work for multi-document summarization \cite{meansum} built multi-source input representations through a simple mean over the last hidden states of the encoder. 
An intrinsic limitation of this method is that the full set of reviews is represented as a single vector.
This aggregation might cause information distortion especially when some input reviews are expected to have conflicted opinions in between.
Standard transformer models \cite{vaswani2017attention} consider only a single input to the decoder part of the model. Aggregating all input reviews into a single input \cite{Junczys-Dowmunt19} with special tokens to represent document boundaries might be slow and impractical due the $O(n^2)$ complexity of the self-attention mechanism. We therefore experiment with several input combination strategies of the transformer cross-attention \cite{LibovickyHM18}.  \\
\textbf{Parallel.} At each cross-attention head, the decoder set of queries $Q$ attend to each of the encoded inputs separately from which the set of keys ($K_i \in K_{1:m}$) and values ($V_i \in V_{1:m}$) are generated and then the yielded context is averaged and followed by a residual connection from the previous decoder layer. 
This corresponds to box (C) in Fig.~\ref{fig:main_figure}.
\begin{align*}
A^h_{\text{parallel}}(Q,K_{1:m},V_{1:m}) = \frac{1}{m}\sum_{i=1}^{m}A^h(Q,K_i,V_i).
\end{align*}
\\
\textbf{Mean.} We also propose a simpler input combination strategy, which is less computationally demanding.
It does not apply the cross-attention with each encoder separately. Instead, the set of keys and values coming from each input encoder are aggregated using the average at each absolute position. 
Afterwards the decoder set of queries attend to this aggregated set of keys and values. This combination can be seen as a more efficient variation of the flat combination strategy~\cite{LibovickyHM18} with mean instead of concatenation. 
Fig.~\ref{fig:meanattention} depicts this strategy, which replaces box (C) in Fig.~\ref{fig:main_figure}.
\begin{align*}
    A^h_{\text{mean}}(Q,K_{1:m},V_{1:m}) = A^h\left(Q,\frac{1}{|m|}\sum_{i=1}^{m}K_i, \frac{1}{|m|}\sum_{i=1}^{m}V_i\right).
\end{align*}
In Sect.~\ref{sect:results}, we compare both approaches through an ablation study, focusing on summary quality as well as empirical training times.
%\todo{needs better explanation}
%
%%%%%%%%%%%%%%%%%%%%%%%%%%%%%%%%%%%%
\begin{table*}[ht]
% add table here with all 
% https://github.com/naverlabseurope/gurkensalad/issues/3#issuecomment-568286696

\resizebox{\textwidth}{!}{
\begin{tabular}{@{}r|l|cccccc@{}}
\toprule
&\textbf{Model} & \textbf{ROUGE-1}  & \textbf{ROUGE-2}  &\textbf{ROUGE-L}  & $\textbf{F}_{\textsc{BERT}}$ & \textbf{Sentiment Acc.} & $\textbf{F}_{\text{category}}$ \\
\midrule
%%%%%%%%%%%%%%%%%%%%%%%%%%%%%%%%%%%%%%%%%%%%%%%%%%%%%%%%%%%%%%%%%%%%%%%%%%%
\parbox[c]{2mm}{\multirow{6}{*}{\rotatebox[origin=c]{90}{YELP}}} 
%& Textrank                  & 28.34     & 4.202      &14.89   & 84.12    & 82.0  & 53.435  \\
%& Lextrank                  & 27.414    & 3.892      &14.89   & 84.174   & 83.5  & 54.132 \\
%& Opinosis                  & 26.808    & 3.419     &14.193  & 81.201   & 80.5  & 53.035 \\
%& H-VAE  & 29,47     & 5.26      & 18.09  & --       & --    & -- \\
%& Meansum                   & 28.64     & 3.821     &15.899  & 86.504   & 83.5  & 50.287  \\
%& Ours                      & \textbf{32.767}    & \textbf{8.658}     & \textbf{18.82}  & \textbf{86.777}   & \textbf{83.92} & \textbf{55.238}   \\
& Textrank \cite{mihalcea-tarau-2004-textrank}  & 28.3     & 4.2      & 14.9   & 84.1   & 82.0  & 53.4  \\
& Lexrank  \cite{RADEV2004919}                  & 27.4     & 3.9      & 14.9   & 84.2   & 83.5  & 54.1 \\
& Opinosis  \cite{ganesan-etal-2010-opinosis}   & 26.8     & 3.4      & 14.2   & 81.2   & 80.5  & 53.0 \\
& H-VAE \cite{brazinskas_unsupervised_2019}     & 29.5     & 5.3      & 18.1   & --     & --    & -- \\
& Meansum  \cite{meansum}                       & 28.6     & 3.8      & 15.9   & 86.5   & 83.5  & 50.3  \\
& Ours                      & \textbf{32.8}    & \textbf{8.7}   & \textbf{18.8}  & \textbf{86.8}   & \textbf{83.9} & \textbf{55.2}   \\
\midrule\midrule
%%%%%%%%%%%%%%%%%%%%%%%%%%%%%%%%%%%%%%%%%%%%%%%%%%%%%%%%%%%%%%%%%%%%%%%%%%%
\parbox[c]{2mm}{\multirow{4}{*}{\rotatebox[origin=c]{90}{RT}}}
%& Textrank  & 18.98     & 4.338 & 19.409 & 85.293 & \textbf{75.752}  & 41.64   \\
%& Lexrank   & 17.604    & 3.546 & 18.224 & \textbf{85.343} & 73.246  & 40.949  \\
%& Opinosis  & 15.191    & 2.931 & 16.873 & 84.086 & 67.535  & 37.162  \\
%& Ours      & \textbf{20.875}    & \textbf{4.511} & \textbf{22.67}  & 85.3   & 70.942  & \textbf{43.612}  \\
& Textrank  & 19.0    & 4.3 & 19.4 & \textbf{85.3} & \textbf{75.8}  & 41.6   \\
& Lexrank   & 17.6    & 3.5 & 18.2 & \textbf{85.3} & 73.2  & 40.9  \\
& Opinosis  & 15.2    & 2.9 & 16.9 & 84.1 & 67.5  & 37.1  \\
& Ours      & \textbf{20.9}    & \textbf{4.5} & \textbf{22.7}  & \textbf{85.3}   & 70.9  & \textbf{43.6}  \\
%%%%%%%%%%%%%%%%%%%%%%%%%%%%%%%%%%%%%%%%%%%%%%%%%%%%%%%%%
\bottomrule
\end{tabular}}
  % automatic evaluation 
\caption{\footnotesize{Automatic evaluations results against gold summaries of Yelp and Rotten Tomatoes (RT) datasets. ``Ours'' denotes our proposed system with parallel input combination strategy and control codes.}}
%Meansum \cite{meansum}, H-VAE is the  \cite{brazinskas_unsupervised_2019}.
\label{table:quantitvative_results}
\end{table*}

\begin{table*}[ht]
% referenceless evaluation 

\centering
\footnotesize
\begin{tabular}{@{}r|l|ccc|ccc@{}}
\toprule
&\textbf{Model} & Dist-1  & Dist-2  & Dist-3  & Dist$_{c}$-1  & Dist$_{c}$-2   &Dist$_{c}$-3 \\
\midrule
%%%%%%%%%%%%%%%%%%%%%%%%%%%%%%%%%%%%%%%%%%%%%%%%%%%%%%%%%%%%%%%%%%%%%%%%%%%
\parbox[t]{2mm}{\multirow{3}{*}{\rotatebox[origin=c]{90}{\footnotesize{Extract.}}}}
& Textrank                  & 0.68     & 0.95      & 0.992     &  0.135     & 0.62  & 0.90  \\
& Lextrank                  & 0.70     & 0.96      & 0.994     & 0.144     & 0.6  & 0.92 \\
& Opinosis                  & 0.72     & 0.94     & 0.97     & \textbf{0.159}    & \textbf{0.66} & \textbf{0.92} \\
\midrule
\parbox[b]{2mm}{\multirow{2}{*}{\rotatebox[origin=c]{90}{\footnotesize{Abstr.}}}} & Meansum                   & 0.72     & 0.95     & 0.98     & 0.091    & 0.39  & 0.67  \\
& Ours                      & \textbf{0.79}    & \textbf{0.99}   & \textbf{1.00}  & 0.097   & 0.41 & 0.64   \\
%%%%%%%%%%%%%%%%%%%%%%%%%%%%%%%%%%%%%%%%%%%%%%%%%%%%%%%%%%%%%%%%%%%%%%%%%%%
\bottomrule
\end{tabular}  % referenceless eval. 
\caption{\footnotesize{Referenceless evaluation results on Yelp dataset.}}
\label{table:referenceless_eval}
\end{table*}

\begin{table*}[ht]
% should be ablation text 

% add table here with all 
% https://github.com/naverlabseurope/gurkensalad/issues/3#issuecomment-568286696

\centering
\footnotesize
\begin{tabular}{@{}l|cccccc|c@{}}
\toprule
& \multicolumn{6}{c}{\textit{Quality}} & \multicolumn{1}{c}{\textit{Speed}} \\
\textbf{Model} & \textbf{ROUGE-1}  & \textbf{ROUGE-2}  &\textbf{ROUGE-L}  & \textbf{F}$_{\text{BERT}}$ & \textbf{Sentiment Acc.} & \textbf{F}$_{\text{category}}$ & \textbf{Train.} (wps) \\
\midrule
%%%%%%%%%%%%%%%%%%%%%%%%%%%%%%%%%%%%%%%%%%%%%%%%%%%%%%%%%%%%%%%%%%%%%%%%%%%
%Ours - \footnotesize{all} & 26.24  & 3.967 &  15.653	& 84.749 & 68.0 & 41.896\\
%Ours - \footnotesize{inferred codes}    & 25.284 & 3.653 &  15.522	& 85.233 & 76.884	& 43.927 \\
%Ours               & \textbf{32.767}    & \textbf{8.658}     & \textbf{18.82}  & \textbf{86.777}      & \textbf{83.92} & \textbf{55.238}   \\
Ours$_{\text{Parallel}}$ & \textbf{32.8}   & \textbf{8.7}     & \textbf{18.8}  & 86.8      & \textbf{83.9} & 55.2   & 3785 \\
Ours$_{\text{Mean}}$ & 29.4 & 5.3 & 17.2 & \textbf{87.6} & 83.4 & \textbf{56.2} & 8075\\
% Ours_{Parallel}-\footnotesize{all} & 26.2  & 4.0 &  15.7	& 84.7 & 68.0 & 41.9 & & \\
Ours$_{\text{Parallel}}$  $-$  cntrl.    & 25.3 & 3.7 &  15.5	& 85.2 & 76.9	& 43.9   & 7609 \\
Ours$_{\text{Mean}}$  $-$  cntrl. & 27.5 & 5.3 & 17.1 & 87.3 & 80.0 & 52.1 & \textbf{8714}\\
\bottomrule
\end{tabular}

\caption{\footnotesize{Ablation study showing the effectiveness of parallel-cross attention and control tokens on Yelp dataset. ``$- \text{cntrl.}$'' denotes models trained without the control step. ``Train. (wps)'' denotes the word per second rate at training time.}}
\label{table:ablation}
\end{table*}

\section{Experimental Setup}
%%%%%%%%%%%%%%%%%%%%%%%%%%%%%
%%%%% EXPERIMENTS SETUP %%%%%
%%%%%%%%%%%%%%%%%%%%%%%%%%%%%
\label{sect:experiments}
\paragraph{Experimental Details}
All our models are implemented with PyTorch~\cite{NIPS2019_9015} and Fairseq~\cite{ott2019fairseq} libraries, as well as scikit-learn~\cite{scikit-learn} for the classifiers used either for inferring control tokens or for evaluation.
For all our models we use sentence piece~\cite{bpe} as a tokenizer with a vocabulary size of \numprint{32000}.
We use the same hyperparameters as the Transformer Big model described by \citet{vaswani2017attention} ($d_{\text{model}}=1024$, $n_{\text{heads}}=16$, $n_{\text{layer}}=6$, $\text{dropout}=0.1$). 
We optimize them with a Nesterov accelerated SGD optimizer with a learning rate of $0.01$.
We train all models for a total of \numprint{80000} steps across $25$ epochs, with linear warm-up for the first \numprint{8000} steps.
We select the best model checkpoint based on perplexity on the validation set.
All models were trained on one machine with 4 NVIDIA V100 GPUs, the longest model took~$50$ hours to train. 
For inference, we use a beam size of~35. We discard hypotheses that contains twice the same trigram.
We limit generation of each summary to a maximum budget of~150 tokens for each summary for Yelp, as was done by \citet{meansum}, and a budget of~50 tokens for Rotten Tomatoes. We set a similar budget for all other extractive baselines in the experiments.
Finally, we use length normalization \cite{DBLP:journals/corr/WuSCLNMKCGMKSJL16} with length penalty 1.2 to account for the model's bias towards shorter sequences.
\paragraph{Datasets}
We evaluate our proposal on two English  datasets: Yelp\footnote{\url{https://www.yelp.com/dataset/challenge}} \cite{meansum} and Rotten Tomatoes \cite{wang2016}.
The Yelp  dataset contains reviews of businesses (approximately one million reviews for around 40k venues). As described in Section~\ref{sect:selfsuper}, for each venue, we select the best reviews to use as target summaries: either the top-$p$ (with $p=15\%$) or the top-$T$ (with $T=100$) reviews, whichever is smaller.
For each target summary thus selected, we then take its~$k=8$ most similar reviews (cosine similarity) to form its input.
We obtain around 340k training examples, representing 22.5k venues. \\
The Rotten Tomatoes dataset was constructed by \cite{wang2016} from the movie review website \url{rottentomatoes.com}.
We use the same process as for Yelp, but use $p=1\%$ and $T=150$.
We construct around 170k training examples, representing 3.7k movies. 
Details about dataset sizes and splits are in the Appendix. %~\ref{appendix:dataset} 
\paragraph{Evaluation Metrics}
We evaluate summary systems with the classical ROUGE-F-\{1,2,L\} metrics~\citep{lin-2004-rouge}.\footnote{For Yelp we use the MeanSum~\cite{meansum} implementation to keep results comparable while for RottenTomatoes we use \textit{py-rouge} package~\url{ pypi.python.org/pypi/pyrouge/0.1.3} }
We also report BERT-score \citep{bert-score}, a metric that uses pre-trained BERT~\citep{devlin-etal-2019-bert} to compute the semantic similarity between a candidate summary and the gold summary.
$\text{Dist-}n$ and $\text{Dist}_c$-$n$ ($n=1,2,3$) scores~\cite{li2016} are the percentage of distinct $n$-grams in the generated text on the summary level or the corpora level respectively. Dist-$n$ is an indicator of repetitiveness within a single summary while Dist$_{c}$-$n$ indicates the diversity of different generations.    
Finally, as done by \citet{meansum}, we use a classifier to check whether the sentiment of the summary is consistent with the sentiment of input reviews (Sentiment Acc.\ in Table~\ref{table:quantitvative_results}).\footnote{We use a 3-class classification: negative (1 or 2 star), neutral (3), positive (4 and 5)). As a result, the numbers are not comparable with those reported \citet{meansum}.}
We extend this method to check whether the correct product category can also be inferred from the summary, we report $\mathbf{F}_{\text{category}}$ the micro F-score of the multi-label category classifier.
%
%%%%% baselines %%%%%%%%%%
\paragraph{Baselines and Other Systems}
We compare our system to three unsupervised baselines.
TextRank \cite{mihalcea-tarau-2004-textrank} and LexRank \cite{RADEV2004919} are extractive systems based on the PageRank algorithm.
Opinosis \cite{ganesan-etal-2010-opinosis} is an abstractive graph-based system.
We use openly available Python implementations for TextRank\footnote{\url{https://github.com/summanlp/textrank}} \cite{DBLP:journals/corr/BarriosLAW16} and LexRank.\footnote{\url{https://github.com/crabcamp/lexrank}}
We use the default parameters of the implementations. 
For Opinosis, we use the official Java implementation\footnote{\url{https://github.com/kavgan/opinosis-summarization}}
with default hyperparameters.\footnote{Except for the redundancy parameter which was set to one, since the default led to many empty outputs.}
%For all baselines, we set a budget of 100~tokens per summary for Yelp and 20~tokens per summary for Rotten Tomatoes.

We also compare our systems with more recent neural unsupervised summarization systems.
For the Yelp dataset, we rerun the released pretrained version of MeanSum\footnote{\url{https://github.com/sosuperic/MeanSum/}} \cite{meansum}.

%%%%%% EXAMPLES %%%%%%%%%%%

\begin{figure}[t]
\resizebox{\columnwidth}{!}{
    \scriptsize	
    \input{figures/examples.tex}
}
\caption{Examples of different model generations to the same input set of documents. Green (italics) denotes substrings with exact match with the input, red (underlined) denotes statements without support in the input. TextRank is shown as a reference: all substrings are present in the input, but note the lack of cohesion.}
\label{fig:examples}
\end{figure}

\begin{figure}[h!]
\resizebox{\columnwidth}{!}{
\input{figures/examples_control.tex}
}
\caption{Examples of outputs summaries \textbf{generated from the same input} when different ``correct'' and ``incorrect'' control tokens are fed as prefixes at inference time.
Control tokens that occur in the summary are highlighted (green for the first rows, red for the other two).}
\label{fig:control_eval_examples}
\end{figure}
%
%%%%%%%%%%%%%%%%%%%%%%%%%%%
%
\begin{figure}[h!]
\centering
    \resizebox{\columnwidth}{!}{
        \includegraphics{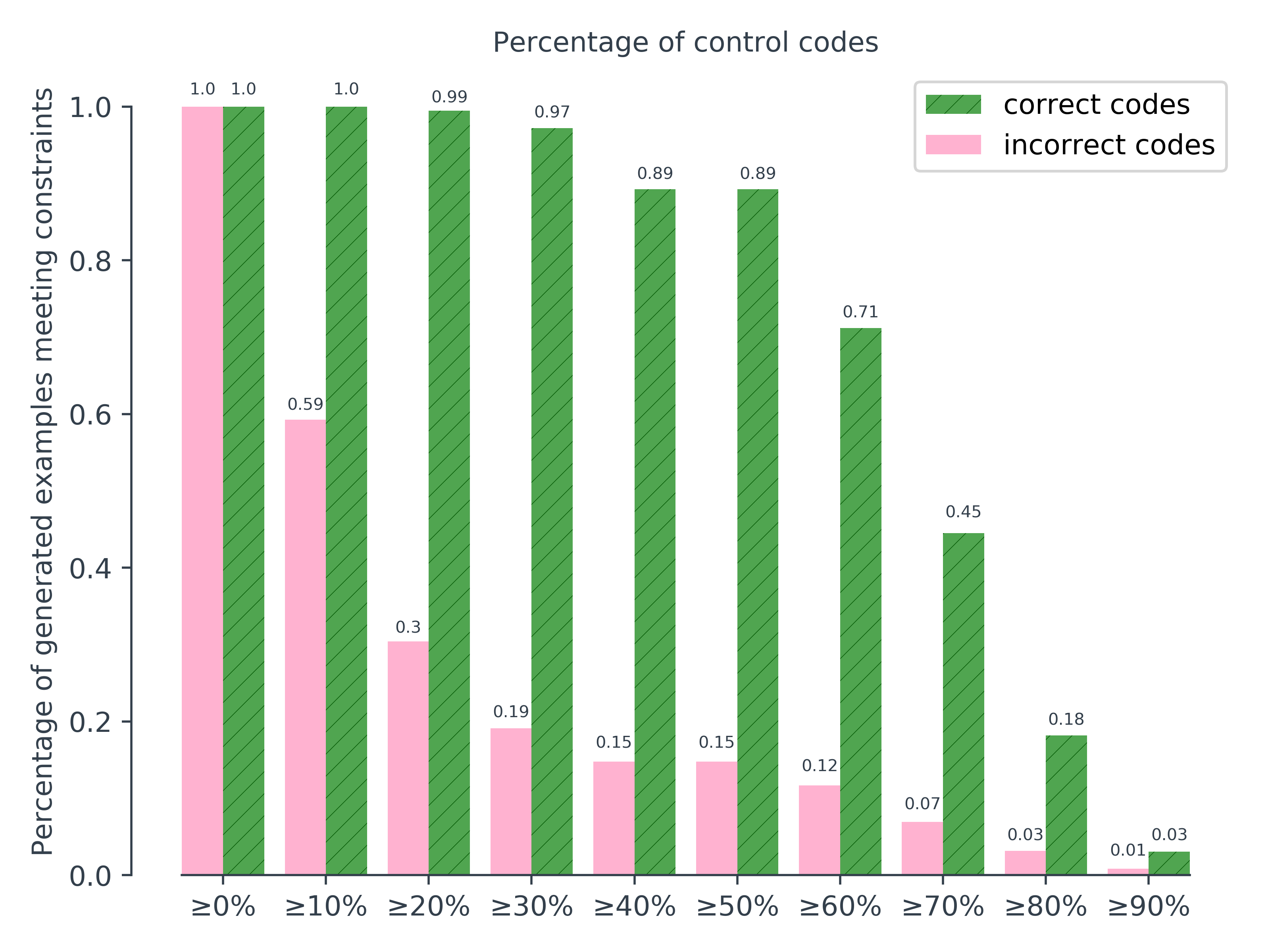}
    }
    \caption{Proportion of control tokens fed as prompts that occur in the generated summary, for the setting of \textit{correct} and \textit{incorrect} control tokens.
    When the model is fed control tokens that occur in the input reviews (\textit{correct}) it tends to generate them in the output.
    This is not the case when it is fed \textit{incorrect} control tokens: it mostly ignores them.}
    \label{fig:control_eval}
\end{figure}

%%%%%%%%%%%%%%%%%%%%%%%%%%%%%%%%%%%%%%%%%%%%%
%%%%%%%%%%%%%%% RESULTS %%%%%%%%%%%%%%%%%%%%%
%%%%%%%%%%%%%%%%%%%%%%%%%%%%%%%%%%%%%%%%%%%%%

\section{Evaluation Results}
\label{sect:results}

%\subsection{Automatic Evaluation}
\paragraph{Automatic Evaluation}
Table~\ref{table:quantitvative_results} contains the automatic evaluation metrics with respect to reference summaries.
The proposed multi-input self-supervised model with control codes perform consistently better in the Yelp dataset across the benchmarked models, inlcuding the recent neural unuspervised models of MeanSum and H-VAE.
Note that because of the concurrent nature of the \citet{brazinskas_unsupervised_2019} paper, the H-VAE model is not available and we report the numbers from their paper.\footnote{While the ROUGE implementation might be different, the numbers of the common baselines are very close.}
For MeanSum we re-run their provided checkpoint and run evaluation through the same pipeline.
The BERTScore~\citep{bert-score} differences are closer and seem to favour neural models.

With the RottenTomatoes dataset we only benchmarked the graph-based unsupervised methods, since the released pretrained MeanSum model does not cover the domain of movie reviews.
We attribute the lower score in sentiment accuracy to the fact that the ``summaries'' in RottenTomatoes are critical reviews, written in a very different style than the original reviews.

\smallskip
Table~\ref{table:referenceless_eval} contains reference-less evaluation, analyzing the number of distinct $n$-grams (an indicator of repetitiveness) on the summary level and corpora level.
On the summary level our model outperforms all the baselines, meaning, our model is capable of generating more rich and less repetitive summaries. On the level of all generations our model generates text with more diversity than MeanSum.
In general however extractive models tend to have more diversity on the corpus level as they directly copy from each input separately, while abstractive models tend to learn repetitive patterns present in the training set.

Fig.~\ref{fig:examples} shows summaries generated by different models from the same input. 
We notice that our model learned to copy aspects of the input documents such as restaurant names ``Capricotti's'' and menu items ``the Bobbie'', this is possibly attributed to the cross-attention mechanism in our proposed model. More examples are provided in the supplementary material Appendix. %~\ref{appendix:examples}. 
%%%%%%%%%%%%%%%%%%%%%%%%%%%%%%%%%%%%%%%%%%%%%%%%%%%%%
%%%%%%%%%%%%%%%%%%%%%%%%% HUMAN EVAL %%%%%%%%%%%%%%%%
%%%%%%%%%%%%%%%%%%%%%%%%%%%%%%%%%%%%%%%%%%%%%%%%%%%%%

%\subsection{Human Evaluation}
\paragraph{Human Evaluation}
Existing natural language generation systems are known to generate very fluent language, that looks very natural to native speakers.
On the other side, current neural models are known to generate factually incorrect data, something which was less of a concern in pre-neural methods but also much harder to detect.
As mentioned by \citet{kryscinski2019}: ``Neither of the methods explicitly examines the factual consistency of summaries, leaving this important dimension unchecked.''
Inspired by~\citet{falke2019} we decided to focus the human evaluation on those aspects of the summarization evaluation in which existing models risk failing the most, the one of \textit{faithfulness}.

We annotated 94 summaries through a crowd-sourcing platform, comparing 3 systems (Gold, MeanSum and ours).
Workers were asked if ``the summary contains correct information given the original reviews''.
In total we had 282 tasks ($94 \times 3$) and each task was labeled by 3 annotators and paid \$0.50 (defined by a pilot study to aim for \$15 / hour) and restricted to experienced, English-speaking workers.
A full description of the campaign, including the filtering of the annotations, is detailed in Appendix. %~\ref{sect:appendixAMT}.

\begin{table}[H]
\centering
\begin{tabular}{l|ccc}
 \toprule
 \textbf{Faithfulness} & \textbf{Gold} & \textbf{Ours} & \textbf{Meansum} \\
 \midrule
 Correct & 67 & 50 & 47 \\
 Incorrect & 3 & 4 & 12 \\
 \%Correct & 95.71 & 92.59 & 79.66 \\
 \bottomrule
\end{tabular}

\caption{Results of the human evaluation focused on faithfulness of generated reviews.}
\label{table:humaneval}
\end{table}
The results in Table~\ref{table:humaneval} show that 92.6\% of the generated summaries of our system are considered factually correct (compare with 95.7\% for the gold summaries), as opposed to 79.7\% of MeanSum.
\paragraph{Ablation}
%\label{sec:ablation}
We analyzed the impact of our proposed variations of the basic self-supervised setting in Table~\ref{table:ablation}.
Removing control codes degrades significantly -- as expected -- sentiment and category classification of the produced summary $F_1$.
It also impacts greatly on the ROUGE score.
Changing the decoder-encoder attention from parallel to mean (see Sect.~\ref{sect:transformer-model}) also degrades ROUGE.
The difference of this attention change without control codes is smaller but -- surprisingly -- in the different direction.

\paragraph{Control Codes}
%\label{sect:incorrectcodes}
The previous ablation study shows the importance of the control codes in the quality of the final summaries.
In order to see how rigidly the model follows those control codes we devise the following experiment to see if the tokens used as control codes are forced to appear in the output text, independent of the input text.

For this, we sample $500$ reviews (for $279$ venues from the Yelp validation set).
For each input example, we randomly sample~$8$ control tokens (inferred control codes, see Sect~\ref{sect:controlcodes}) from the tokens occurring in the review.
We refer to these as \textit{correct control tokens}. 
We run the decoder using these control tokens as prompt and count the proportion of them that also occurs in the generated summary.
For comparison, we repeat the same experiment but sampling instead $8$ control tokens that do \textit{not} occur in the input text.
We refer to these as \textit{incorrect control tokens}. 

To minimize the possibility of conditioning on control tokens that might show up naturally in the generated text, for both settings, we repeat the process~$5$ times per input example (resulting in $2500$ with \textit{correct control tokens} as prefix and $2500$ using \textit{incorrect}).
We report in Fig.~\ref{fig:control_eval} the proportion of fed control codes that are generated by the model in both cases.
We observe that the model tends to comply with the correct control tokens that occur in the input documents (eg: $89\%$ of the summaries contain more than $50\%$ of the control tokens), but tends to ignore the control tokens when they do not occur in the input.
Fig.~\ref{fig:control_eval_examples} shows a set of generated examples for the same input when the model is conditioned on different control tokens.

\section{Conclusion}

Neural methods have shown great promises for unsupervised multi-document abstractive summarization, overcoming the lack of fluency of extractive models.
However, those models are often complex to train and more importantly tend to generate incorrect statements; characteristics which are exacerbated in the unsupervised setting.
Our proposed models aim to overcome those problems by proposing a simple training mechanism relying on a self-supervised formulation.
In addition to our use of multi-input transformers and control codes, we show that the resulting summaries are better (as measured by ROUGE and other automatic measures), and produce more faithful summaries (as measured by human evaluation).
The use of control codes makes it easy to extend for other multi-document summarization use-cases.

While the generated reviews are more factual than those generated by other models, we want to stress that inaccuracies can still appear and that special care should be taken if such methods are to be deployed.
In particular, the models learn the conjugations from the input, which is mostly in first persons.
Such summaries might be misleading as it could lend to believe that an actual human wrote those.
We recommend strongly that any use of such algorithms to be accompanied by a clear disclaimer on its true nature.

\bibliography{main}
\bibliographystyle{icml2020}

\twocolumn[
\icmltitle{Supplementary Material: Self-Supervised and Controlled Multi-Document Opinion Summarization}
% \icmlsetsymbol{equal}{*}
% \begin{icmlauthorlist}
% \icmlauthor{Aeiau Zzzz}{equal,to}
% \end{icmlauthorlist}
% \icmlaffiliation{to}{Department of Computation, University of Torontoland, Torontoland, Canada}
% % the "keywords" metadata in the PDF but will not be shown in the document
% \icmlkeywords{Machine Learning, ICML}
\vskip 0.3in
]
\appendix

\pagebreak

\begin{table*}[]
\centering
\begin{tabular}{@{}lccc@{}}
\toprule
Dataset                & \multicolumn{2}{c}{Reviews} & Businesses / Movies \\ 
                       & Train         & Valid       &                     \\\midrule
Yelp | Ours            & 349,839       & 48,677      & 22,522              \\
Yelp | Ours + Control  & 404,811       & 47,938      & 22,522              \\
Rotten Tomatoes | Ours & 167,168       & 18,731      & 3,732               \\ \bottomrule
\end{tabular}
    \caption{Sizes of Training and validation splits of different datasets. }%
    \label{fig:appendix_datasetsizes}%
\end{table*}

\section{Generated Examples}
Fig.~\ref{fig:AppendixExamples},\ref{fig:AppendixExamples2} include a set of samples generated from our model and baselines, full generations our model can be downloaded~\url{https://www.dropbox.com/s/w6eqviy5fnda11f/hypos_and_refs.zip?dl=0}
% add examples from here :https://docs.google.com/document/d/1Rq9emhK-GbF51RlGrHlEkAND-SxAtQpIMlOo4R3X3j4/edit

\begin{figure*}[t]
\footnotesize
\noindent\fbox{%
    \parbox{\linewidth}{%

\textbf{Inputs:}
\\
\textbf{1.} Best Philly Ever!!! Thank You Sam!!! Sometimes it is the little things in life that can Make You Happy- All it took was a Perfect Cheese Steak to Cheer Me Up, not to mention seeing a Friend Again - Thanks again Sam,, It wouldn't be the same without You 
\\
\textbf{2.} Wow after all the hype about what a great place I was really disappointed. If this is a franchised operation than the quality control is really lacking. Our first visit to Capriotti's and with so many other quality places I doubt if they will get us as repeat customers. Well, here it is. We ordered the Bobbie and the Capastrami shared it. Both had cold bread in fact we got the impression that both sandwiches had been pre made and put in a refrigerator because the insides were also cold. No taste at all in either. For a company that supposedly cooks overnight you would think the turkey ingredients would look like turkey but apparently they shred it into little tiny bits. Will not return 
\\
\textbf{3.} This place is always good, I think the owner actually made my sandwich last time I was there , owner or manager, anyway it was superb! quite flavorful, even the next day it tasted just as good. Grab a Capistrami you can't go wrong, until next time Cappie's , be well. 
\\
\textbf{4.} one New Year's resolution is to write more Yelp reviews, so here goes... In Vegas for NYE and gave this place a shot per other Yelp reviews. I had the Capistrami and the girlfriend had the Cheese Steak, which I had few bites of. Both were absolutely delicious in an awesome-deli-sandwich-sort-of-way. The shop is no-frills with only some bar seating, but the sandwiches are really reasonably priced. So if all you're after is a fantastic deli sandwich, definitely go. 
\\
\textbf{5.} number 1 in Vegas for a reason. Everyone has their favs.... the capistrami, the cheese steak with mush... Mine is definitely the Bobby. In case you haven't viewed their menu yet, the bobby is thanksgiving leftovers in a huge sandwich... yeah, exactly. 
\\
\textbf{6.} Worst service I have seen at a capriotti's. Ordered a sandwich, watched the guy write it down and 25 minutes later the same person asked what I wanted when I reminded him of my sandwich, he only said he can't remember where the order went. I watched 4 people come in after me order, one person the same sandwich just a different size then me get their food, pay and leave. I will not be coming back to the location ever again. Looks like I will be going to firehouse for now on! 
\\
\textbf{7.} Stopped in for a sandwich on the way to the park. Next day I notice the charge has had a \$2 tip added to it that I did not authorize. ( I left a cash tip in the beer money jar) I called Corporate and got nowhere because this is a franchise store. At that point I gave up because as much as I like their sandwiches I am never going back. 
\\
\textbf{8.} Don't bother calling in an order. If they tell you a time it will be off by at least thirty minutes. Terrible service. Great food tho.
\\

\textbf{Summary OURS:}\\
This was my \g{first visit} to \g{Capriotti's} and I really enjoyed it . I had the \g{Capastrami} and my \r{husband} had \g{the Bobbie} . We both enjoyed our \g{sandwiches} as well . The \g{quality} of the \g{ingredients}, however, was not what we expected . We also enjoyed the \g{cheese steak} as well as the \g{turkey}, which was not bad at all . This place is a bit on the \r{expensive} side for what you get, but you get what you pay for . The \g{seating} is limited, so it's a good place to visit if you're in a hurry.
\\

\textbf{Summary Meansum:}\\
Drove by here for the \g{first time}. I just went to the \g{deli} with a friend and it's a quick fix that is just about as good as it gets. But it's not an actual \g{sandwich}, but it's \r{not as good as I remembered it}, but they were great!! \g{Sandwich} was also very good, just a \r{hint of cinnamon}. I will be back for the other locations. 
\\

\textbf{Summary TextRank (Extractive):}\\
Will not return  This place is always good, I think the owner actually made my sandwich last time I was there , owner or manager, anyway it was superb! Ordered a sandwich, watched the guy write it down and 25 minutes later the same person asked what I wanted when I reminded him of my sandwich, he only said he can't remember where the order went. I watched 4 people come in after me order, one person the same sandwich just a different size then me get their food, pay and leave. At that point I gave up because as much as I like their sandwiches I am never going back.
    }%
}

\caption{Examples of output summaries for different models.}
\label{fig:AppendixExamples}
\end{figure*}

\begin{figure*}[t]
\footnotesize
\noindent\fbox{%
    \parbox{\linewidth}{%
\textbf{Inputs:}
\\
\textbf{1.} Great service and a super clean nice location here. Considering this is in a busy airport, I was impressed. The pricing here, which is about double to triple regular prices, was what knocks off a star for me. 
\\
\textbf{2.} If you're by the D gates at Sky Harbor, this is your coffee stop. Much better than the Starsucks at the high C gates. 
\\
\textbf{3.} Spotted in due to flight delay. The big comfy brown lawyer seats is what attracted me in. I figured I could get some work done. I ordered a non fat vanilla iced tea. It was pretty good. I noticed the prices were a lil bit more expensive. 
\\
\textbf{4.} \$9 for a sandwich. I guess you can charge whatever you want when the airport doesn't have any other options in concourse C 
\\
\textbf{5.} The line is quick, the people are friendly and the drinks are tasty. Also for skyharbor employees, they actually give an airport discount, unlike Starbucks. 
\\
\textbf{6.} Try gingerbread latte yum. This is the best looking, most comfortable airport coffee shop I've ever been in !!! Big comfy chairs with little tables. A big water container with cups in the restaurant away from the congestion of the order counter. Friendly , happy workers equals happy customers. I know you can't please everyone, but at 5am a room full of happy airline travelers is a hard thing to come by. If your in need of coffee in Phoenix Sky Harbor , terminal 3 be sure to stop by and take a load off !! 
\\
\textbf{7.} The lid fell off my cup, burned my hand, and spilled half of my coffee. Employees never asked if I was okay, or offered to replace my coffee. Will not be back to this location. 
\\
\textbf{8.} Delicious cup of coffee. Very impressed Mr. Peets. Will be returning whenever I can.
\\

\textbf{OURS:}\\
Try the \g{gingerbread} cups. The coffee and ginger cups are \g{delicious} and the \g{chairs are comfortable} . I've been working in the \g{terminal} for a long time . This is a must stop if you're in the \g{airport} . The staff is \g{friendly} . 
\\

\textbf{Meansum:}\\
\r{5.50 for a 2.5" breakfast sandwich}. \r{I'm a big fan of the concept} but this place is way better than \g{Starbucks}. The staff is \g{friendly}, and fast. \r{I'm not a big fan of sweets} but I'd be happy to come back.
\\

\textbf{TextRank (Extractive):}\\
Great service and a super clean nice location here. Considering this is in a busy airport, I was impressed. The pricing here, which is about double to triple regular prices, was what knocks off a star for me. I noticed the prices were a lil bit more expensive. I guess you can charge whatever you want when the airport doesn't have any other options in concourse C   The line is quick, the people are friendly and the drinks are tasty. This is the best looking, most comfortable airport coffee shop I've ever been in !!! Will not be back to this location.
    }%
}
\caption{Examples of output summaries for different models.}
\label{fig:AppendixExamples2}
\end{figure*}

\pagebreak
\section{Inferred Control Tokens}
\label{appendix:examples}
Fig.~\ref{fig:appendixAspects} shows examples of the top inferred tokens for some categories in the Yelp dataset, those tokens have been inferred using our proposed method in this work.  
\begin{figure*}[t]
\footnotesize
\noindent\fbox{%
\parbox{\linewidth}{%
\textbf{Delis: }
deli, sandwiches, sandwich, bagels, skinnyfats, subs, bagel, sub, chompie, smoked meat
\\
\textbf{Nail Salons: }
nails, pedicure, nail, pedicures, pedi, salon, manicure, pedis, colors, salons
\\
\textbf{Sushi Bars: }
sushi, hibachi, kona, rolls, roll, japanese, ayce, sake, benihana, poke
\\
\textbf{Florists: }
flowers, trader, arrangement, florist, wedding, bouquet, tj, arrangements, aj, grocery
\\
\textbf{Beauty \& Spas: }
walgreens, tattoo, sephora, ti, vdara, tattoos, haircut, barbers, barber
\\
\textbf{Party \& Event Planning: }
herb box, wedding, kids, fun, party, event, golf, painting, rainforest, blast
\\
\textbf{Trainers: }
gym, workout, fitness, equipment, membership, trainers, training, trainer, instructors, machines
\\
\textbf{Cafes: }
cafe, first watch, bouchon, salsa bar, café, coffee, breakfast, gallo, crepes, latte
\\
\textbf{Mags: }
books, store, book, games, bookstore, selection, records, comics, vinyl, game
\\
\textbf{Ice Cream \& Frozen Yogurt: }
gelato, ice, sonic, yogurt, custard, culver, flavors, freddy, froyo, icecream
\\
\textbf{Burgers: }
burgers, burger, mcdonald, ihop, applebee, red robin, mcdonalds, wellington, hamburgers, castle
\\
\textbf{Furniture Stores: }
furniture, ikea, mattress, store, sales, delivery, bought, couch, purchase, bed
\\
\textbf{Sporting Goods: }
bike, bikes, shoes, gear, gun, store, range, golf, shop, equipment
\\
\textbf{Bakeries: }
bakery, pastries, wildflower, cupcakes, panera, cake, pastry, cookies, cinnamon rolls, cakes
\\
\textbf{Thai: }
thai, curry, pad, asian, khao, curries, food, papaya, satay, tom
\\
\textbf{Gyms: }
gym, workout, fitness, membership, equipment, trainer, trainers, work out, coaches, class
\\
\textbf{Cosmetics \& Beauty Supply: }
walgreens, pharmacy, products, haircut, store, sephora, hair, makeup, lashes, kohl
\\
\textbf{Auto Repair: }
car, vehicle, dealership, cars, auto, mechanic, vehicles, oil, windshield, tire
\\
\textbf{Department Stores: }
walmart, target, costco, store, department, shopping, mall, section, ross, sears
\\
\textbf{Local Services: }
post office, thrift, laundromat, daycare, guitar, cleaners, pest, activities, storage, laundry
\\
\textbf{Hair Extensions: }
hair, salon, stylist, color, haircut, extensions, appointment, she, lashes, blow
\\
\textbf{Hair Removal: }
eyebrows, nails, pedi, pedicure, nail, appointment, brows, wax, salon, waxing
\\
\textbf{Laundry Services: }
cleaners, clothes, laundry, cleaning, dry, laundromat, dress, pants, machines, shirts
\\
\textbf{Doctors: }
dr, doctor, doctors, medical, hospital, office, patients, appointment, nurse, clinic
\\
\textbf{Movers: }
move, moving, movers, company, truck, storage, guys, furniture, moved, haul
\\
\textbf{Printing Services: }
printing, print, ups, package, business, fedex, shipping, customer, printed, store
\\
\textbf{Makeup Artists: }
makeup, hair, salon, make up, lashes, stylist, blow, appointment, eyebrows, brows
\\
\textbf{Plumbing: }
plumbing, company, plumber, water, call, called, work, house, job, leak
\\
\textbf{Real Estate Services: }
property, estate, westgate, home, company, process, house, realtor, rent, work with
}}
\caption{Examples of Inferred control tokens for each category of venues for the Yelp dataset.}
\label{fig:appendixAspects}
\end{figure*}

\section{Human Evaluation Campaign}
\label{sect:appendixAMT}
We used Amazon Mechanical Turk to ask~3 ``workers'' to assess if~282 summaries produced by~3 systems (94 from each: ours, gold from human experts and Meansum) aligned correctly with sets of~8 reviews.
Workers had to read the reviews, the summary and answer the question: ``does the summary contain correct information given the original reviews?'' Instructions specified to ``assess the faithfulness of the summary with respect to [the] set of reviews,'' specifically to ``verify that the summary [did] not contain factually incorrect or self-contradicting statements that could not be inferred from what [was] provided in the original reviews.''
Using Mechanical Turk qualification criteria, we asked for the workers: (1) to be located in the United States, Canada or United Kingdom; (2) to have a HIT approval rate higher than 98; (3) to have more than 1000 HITs approved.

We did an internal run to estimate the time needed per individual assignment --each Human Intelligence Task, or HIT, an annotation in our case, was assigned to~3 workers. We followed it by a short pilot to validate the average~2 minutes we had estimated. This is important to establish the rate to pay: 2 minutes translate into 30 potential assignments per hour, we picked \$0.50 to target an average \$15 hourly wage.
Beyond the timing, the pilot was also used as a dry run for the full campaign. Average time to Answer and the theoritical hourly wage is available in Table~\ref{table:AppendixAMT}

By using shuffled gold summaries, hence written for another set of reviews, we included~21 badly aligned ``negatives.'' Workers who answered \textit{yes} for these obvious \textit{no} were filtered out as ``dubious'' from the results: all their answers were discarded. After filtering out the ``negatives'' HITs and the ones from ``dubious'' answers, we were left with~446 annotations. We further discarded all annotations made in less than a minute to keep~377 realistic answers.

Finally we looked for full agreement at the HIT level and kept only the ones with either 0 \textit{yes} or 0 \textit{no}, with varying numbers, from 1 to 3, of the alternatives after the filtering of the "dubious" and "unrealistic" answers.
Not surprisingly, as we focused on alignment, Gold summaries scored best but ours scored nicely, with a very low number of misaligned summaries:

Assessing the alignment of summaries to a set of reviews is not an easy task. We decided to discard all answers from the "dubious" workers who erred on our "negatives" summaries to be on the safe side. Mechanical Turk reports the time taken for an assignment, their averages is an interesting metric to look at, especially the way it evolves along our filterings —we translated it to the associated theoretical hourly wages, alas all under the \$15 we initially targeted.
\begin{table*}[h!]
\centering
\begin{tabular}{@{}l|cccccc|cc@{}}
\toprule
 \textbf{Set} & \textbf{Unfiltered} & \textbf{Negatives discarded} & \textbf{Dubious discarded} & \textbf{less than 1min discarded} & \textbf{Agreement} \\
 \midrule
 \textbf{Average time to Answer} & 2min16s & 2min17 & 2min9 & 2min26 & 2min10 \\
 \textbf{Theoretical hourly wage} & 13.22 & 13.16 & 13.96 & 12.36 & 13.87 \\
 \bottomrule
\end{tabular}
\caption{Average time to Answer and the theoritical hourly wage of turkers (in USD) for the crowdsourcing experiments of human evaluation}
\label{table:AppendixAMT}
\end{table*}

\end{document}